\documentclass{article} 
\usepackage{iclr2023_conference,times}

\usepackage{amsmath,amsfonts,bm}

\def\eqref#1{equation~\ref{#1}}

\def\1{\bm{1}}

\DeclareMathAlphabet{\mathsfit}{\encodingdefault}{\sfdefault}{m}{sl}
\SetMathAlphabet{\mathsfit}{bold}{\encodingdefault}{\sfdefault}{bx}{n}

\usepackage{hyperref}
\usepackage{url}
\usepackage{cleveref}
\usepackage{graphicx}
\usepackage{tikz}
\usepackage{xcolor}
\newcommand*\circled[1]{\tikz[baseline=(char.base)]{
            \node[shape=circle,fill,inner sep=0.2pt] (char) {\textcolor{white}{#1}};}}

\newcommand\nj[1]{\textcolor{black}{#1}}

\title{Analyzing Multimodal Objectives Through the Lens of Generative Diffusion Guidance}  

\author{Chaerin Kong, Nojun Kwak \\
Seoul National University \\
\texttt{\{veztylord,nojunk\}@snu.ac.kr}
}

\iclrfinalcopy 
\begin{document}

\maketitle

\begin{abstract}
Recent years have witnessed astonishing advances in the field of multimodal representation learning, with contrastive learning being the cornerstone for major breakthroughs.
Latest works delivered further improvements by incorporating different objectives such as masked modeling and captioning into the frameworks, but our understanding on \textit{how these objectives facilitate learning} remains vastly incomplete. In this paper, we leverage the fact that classifier-guided diffusion models generate images that reflect the semantic signals provided by the classifier to study the characteristics of multimodal learning objectives. Specifically, we compare contrastive, matching and captioning loss in terms of their semantic signals, and introduce a simple baseline that not only supports our analyses but also improves the quality of generative guidance in a straightforward manner.

\end{abstract}

\section{Introduction}

Vision-Language Pretraining (VLP) has attracted great attention from the community for its wide and robust applications in different downstream tasks. The seminal work of CLIP~\citep{radford2021learning} employs the image-text contrastive objective to successfully embed images and text descriptions in a common feature space, inspiring numerous subsequent works that explore different objectives~\citep{li2022blip, yu2022coca, yang2022vision, jang2023self} and architectures~\citep{li2021align, jang2022unifying, wang2022image}. Recently, cross-modal generative models~\citep{ramesh2021zero, saharia2022photorealistic, rombach2022high, kong2022few} are also gaining wide popularity thanks to the powerful capacity of diffusion models~\citep{sohl2015deep, ho2020denoising} and readily available guidance of vision-language foundation models. These models aim to synthesize or edit images so that the outputs are both realistic and condition-aligned.

Conditional diffusion models embody the conditioning information in two ways: classifier guidance~\citep{dhariwal2021diffusion} and classifier-free guidance~\citep{ho2022classifier}. While classifier-guided models~\citep{nichol2021glide, dhariwal2021diffusion, kong2023leveraging} typically leverage multimodal embeddings of a joint embedding network (\textit{e.g.,} CLIP) using cosine similarities, it is unclear whether this approach is optimal. For example, recent works~\citep{zhong2022regionclip, li2022grounded} have pointed out that global representations of CLIP learned by the contrastive objective are not suitable for handling fine-grained correspondences between image and text as they represent the image and the text \textit{as a whole}. 
This indicates that features learned from sequence level contrastive learning may have certain blind spots that can be made visually apparent when directly applied to the diffusion process as the guidance signal.

In this paper, we aim to obtain a better understanding of different multimodal learning objectives (\textit{e.g., image-text contrastive, image-text matching, image captioning}) by analyzing their semantic signals as generative guidance. That is, we utilize the classifier-guided diffusion process to study the characteristics of different objectives by carefully inspecting the samples they produce, and further present a straightforward modification to the previous method that both supports our findings and improves the generation quality. We note that the aim of this paper is not to present a high-performing generative model. Rather, we are simply \textit{leveraging} the diffusion process to visually analyze different objectives and hypothesize about the properties of the accordingly learned representations.

\section{Multimodal Objective as Generative Guidance}

\vspace{-2mm}
\subsection{Classifier-guided Diffusion}
\vspace{-2mm}

\citet{dhariwal2021diffusion} has introduced classifier guidance as a means to steer the generative diffusion process towards the conditioning class. This requires a \textit{noise-aware} classifier whose gradient can be used to guide the diffusion process. Formally, denoting the predicted parameters of timestep $t$ as $\mu_\theta(x_t)$, $\Sigma_\theta(x_t)$, the next step diffusion sampling becomes
\begin{equation}
    x_{t-1} \sim \mathcal{N}(\mu_\theta(x_t)+s \Sigma_\theta(x_t) \nabla_{x_t}\log p_{\phi}(y|x_t), \Sigma_\theta(x_t)),
\label{eqn:cg1}
\end{equation}
where $s$ is the step size, $y$ indicates the class label and $p_\phi$ refers to the classifier. This formulation was altered by \citet{nichol2021glide} to suit text-to-image generation as follows:
\begin{equation}
    x_{t-1} \sim \mathcal{N}(\mu_\theta(x_t)+s \Sigma_\theta(x_t) \nabla_{x_t} \langle f(x_t),  g(c) \rangle, \Sigma_\theta(x_t)),
\label{eqn:cg2}
\end{equation}
\nj{where $\langle \cdot, \cdot \rangle$ indicates the inner product and} $f$, $g$, $c$ are \nj{the} image encoder, text encoder, and the text condition, respectively. 

Overall, the classifier guidance encourages the model to generate samples \nj{that are well-aligned with the condition} 
according to a predefined metric. Obviously, the choice of this metric affects the final output sample, revealing how each metric (objective) connects the two modalities (\textit{image and text, in our case}) in multiple levels.

\vspace{-2mm}
\subsection{Pretrained Models}
\vspace{-2mm}

As our goal is to study the semantic signals encoded in different objectives, we employ a pretrained diffusion backbone and a pretrained vision-language guidance model for our analysis. 

\noindent
\textbf{Generative Model}
For the image generator, we use a 256$\times$256 unconditional diffusion model pretrained on Imagenet\footnote{https://github.com/openai/guided-diffusion}~\citep{russakovsky2015imagenet}. 
We note that this is \textit{not} a state-of-the-art text-to-image diffusion model. The unconditional nature of this model renders it well-suited for our purpose, as it solely relies on the classifier signal for condition-aware synthesis, making it possible to analyze the encoded semantic information in a disentangled manner. Employing an excessively powerful generator can similarly obfuscate our analysis, as its generative capacity can compensate for the weakness in the guidance signal and mask its blind spots.


\noindent
\textbf{Guidance Model}
Among many available candidates, we choose BLIP~\citep{li2022blip} as our main guidance model\footnote{https://github.com/salesforce/BLIP}. This model is trained on 129M image-text pairs simultaneously optimizing for three objectives: image-text contrastive (\textbf{ITC}), image-text matching (\textbf{ITM}) and image captioning (\textbf{CAP}). The fact that a single model can evaluate these three scores makes it an excellent guidance model for our analysis, as we can safely minimize the compounding effects coming from using different models trained with different dataset, architecture and optimization scheme. 
We adopt the idea from \citet{avrahami2022blended} to first predict the denoised version for guidance signal computation.

\vspace{-2mm}
\subsection{Objectives and Benchmarks}
\vspace{-2mm}

We analyze three commonly used multimodal objectives: \textbf{ITC, ITM} and \textbf{CAP}. We simply replace the classifier logit in \cref{eqn:cg1} with the corresponding loss terms of BLIP (with the sign reversed). For details about the loss computation, please refer to \citet{li2022blip}. 
For systematic evaluation, we mainly follow the benchmark of \citet{saharia2022photorealistic}, namely DrawBench. We further add prompts from COCO~\citep{lin2014microsoft} captions. Qualitative evaluations as well as quantitative comparison\nj{s} based on user study are performed to draw insights.

\section{Analyses}

In this section, we first present our empirical findings and propose a straightforward modification called \textbf{SHIFT} that reflects the insights by employing the coarse-to-fine guidance of \textbf{CAP} and \textbf{ITC}. We acknowledge that using a more powerful model would relieve some of the issues demonstrated in this section as the data scale can compensate for the blind spots, but we believe the findings we present here nevertheless hold true and will come into play in an increasingly challenging problem setting. 


\subsection{Findings}
\label{sec:findings}

\begin{figure}[h!]
\vspace{-3mm}
\centering
  \includegraphics[width=0.82\textwidth]{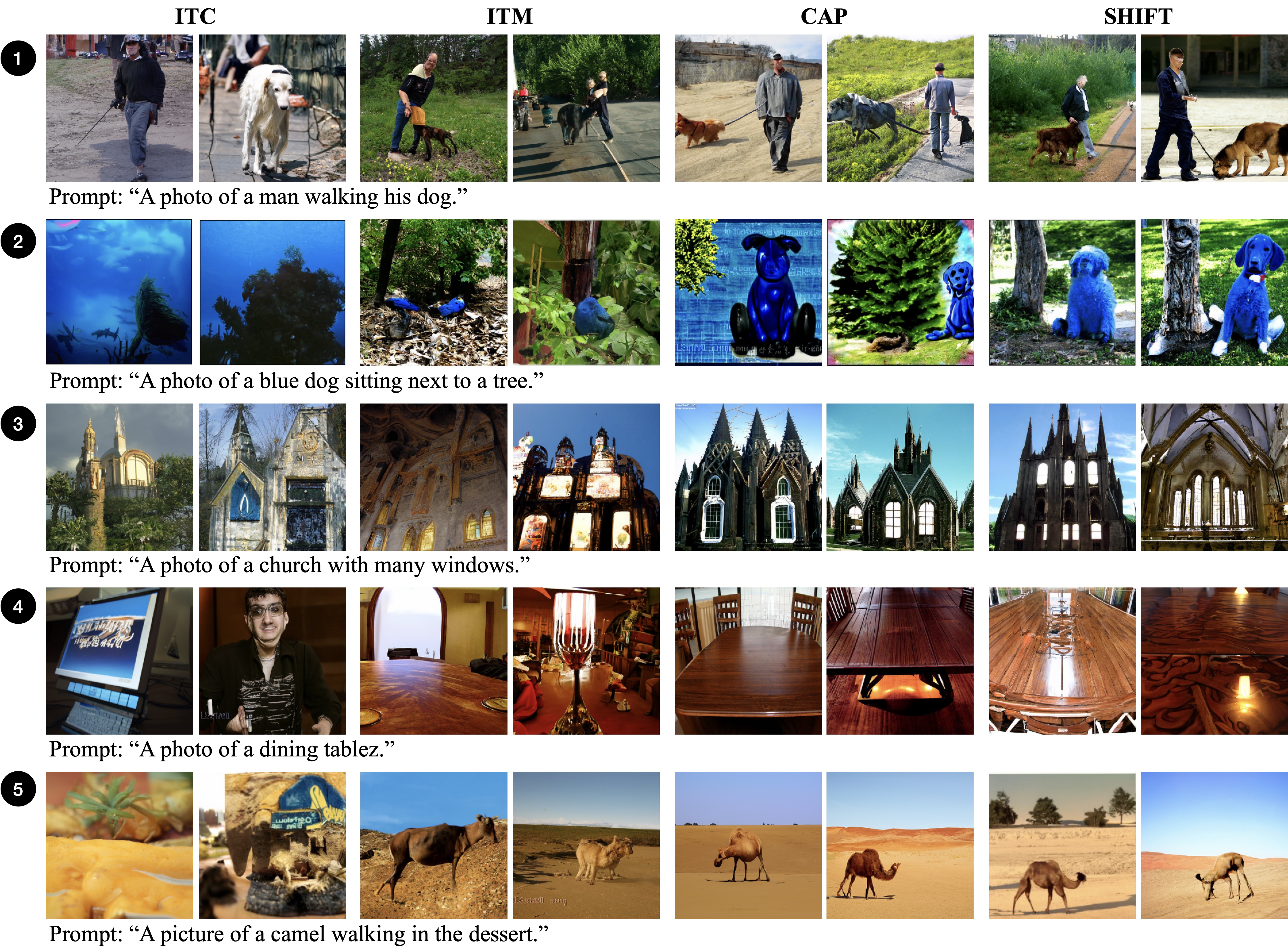}
  \vspace{-3mm}
  \caption{Text-to-image samples using each objective as the classifier-guidance. We only present preliminary examples for illustrative purpose. Please refer to the Appendix for more cases.}
  \label{fig:fig1}
\end{figure}

\textit{1. While \textbf{ITC} focuses on the fine details of the salient object, \textbf{CAP} tends to reason about the global scene composition.}

Looking at \Cref{fig:fig1} row \circled{1}, we can clearly see that contrastive loss is more effective in forming fine details of the main object while it often leaves out \textit{less important} objects or attributes in the given prompt. We hypothesize the former is due to the core dynamic of contrastive learning that aims to learn relative distances by \textit{distinguishing} objects. At the same time, as the objective only compares fully abstracted representations (\textit{i.e.,} \texttt{[cls]} tokens) that dominate each entity as a whole, it fails to densely parse the scene. On the other hand, captioning objective forces to understand the scene structure in a deeper level, making captioning-guided samples more faithful to complex texts. 


\textit{2. \textbf{ITC} commonly fuses visual semantics together to forcefully form a global semantic.}

\circled{2} in \Cref{fig:fig1} illustrate\nj{s} a more extreme case where \textbf{ITC} not only omits semantic components but arbitrarily mix\nj{es} them. From these examples, we can diagnose that the contrastive objective does reflect semantic attributes (\textit{e.g., blue}) but fails to \textit{relate} them to the correct object (\textit{e.g., dog}). This can be another side effect of simplified distance learning. Captioning, in contrast, requires the model to reason about both objects and their relations, deepening the scene-level understanding.



\textit{3. Patch-token cross-attention plays a key role in fine-grained visual understanding.}

We now widen the scope of our analysis by further looking at \textbf{ITM}. In contrary to \textbf{ITC} that simply compares two \texttt{[cls]} tokens, \textbf{ITM} involves lower-level cross-attention between image patch tokens and text tokens (as in \textbf{CAP}) to output a matching score between 0 and 1. We discovered that this operation plays an important role in fine-grained visual understanding and representation robustness. To our surprise, \textbf{ITM}, more or less an auxiliary loss to polish multimodal representations, encodes strong semantic signals that involves dense scene understanding. Looking at \circled{1}, \circled{2}, \circled{3}, cross-attention-based objectives (\textbf{ITM}, \textbf{CAP} and \textbf{SHIFT}) demonstrate capacity for fine-grained visual reasoning, and \textbf{ITM} signal successfully materializes objects with corresponding attributes and relations, though at a lower visual quality.



\textit{4. Dense supervision makes the representations more robust to noise perturbations.}

Last two rows of \Cref{fig:fig1} show the impact of noise in the text prompt. \circled{4} depicts that as opposed to \textbf{ITC} that generates random objects under mild typo, other losses render relatively consistent outputs. \circled{5} delivers a similar insight, where the phrase `camel in the \textit{dessert'} not \textit{`desert'} is likely mistaken by the text provider. These cases are very probable in the typical setting where massive noisy image-text data are crawled from the web, and we observe that dense supervision that involves low-level patch-token cross-attention shows better robustness against textual perturbations, as perturbed text inputs attend to not only themselves but also visual information to form a more robust representation. 


\textit{5. \textbf{CAP} is a more indirect if not challenging form of supervision than \textbf{ITC} or \textbf{ITM}.}

\begin{figure}[h!]
\centering
  \includegraphics[width=0.8\textwidth]{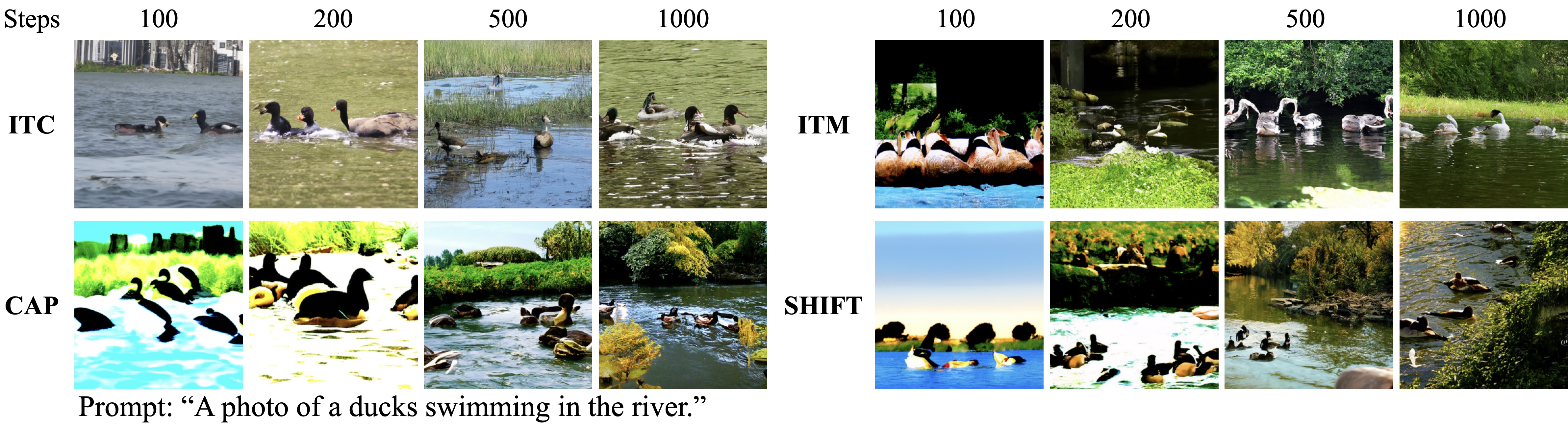}
  \vspace{-3mm}
  \caption{Generated samples for each objective and the number of diffusion sampling steps.}
  \vspace{-2mm}
  \label{fig:fig2}
\end{figure}

Lastly\nj{,} we inspect the optimization complexity of each objective by differing diffusion sampling steps. As each diffusion step corresponds to an update using the loss gradient, we regard an objective that generates reasonable sample with fewer steps to have lower optimization complexity. Referring to \Cref{fig:fig2}, we see that \textbf{ITC} and \textbf{ITM} clearly take less steps to output realistic samples compared to caption-based losses. This observation coincides with \citet{radford2021learning} and \citet{yu2022coca}, where the former explicitly chose contrastive loss for training efficiency and the latter has been reported to take much more resources to converge due to captioning. We conclude that as captioning demands a more semantic visual understanding, learning becomes trickier compared to the simple distance learning.

\subsection{New Baseline: Guidance Shift}

Based on the above findings, we propose a simple yet effective baseline that takes advantage from both ends, \textit{i.e.,} contrastive learning and captioning. To leverage the strengths from both, we introduce guidance shift, where we start with captioning loss and gradually shift to contrastive loss for the generative guidance. Formally, our \textbf{SHIFT} loss can be written as:
\begin{equation}
    \mathcal{L}_{SHIFT} = t\mathcal{L}_{ITC} + (1-t)\mathcal{L}_{CAP},
\end{equation}
where $t$ is the normalized time step, progressing from 0 to 1. The idea is to first outline the overall structure with \textbf{CAP} and then refine the details with \textbf{ITC}. To study its effectiveness, we conduct quantitative user study as well as qualitative evaluations presented in the Appendix. 

\begin{figure}[h!]
\vspace{-1mm}
\centering
  \includegraphics[width=0.76\textwidth]{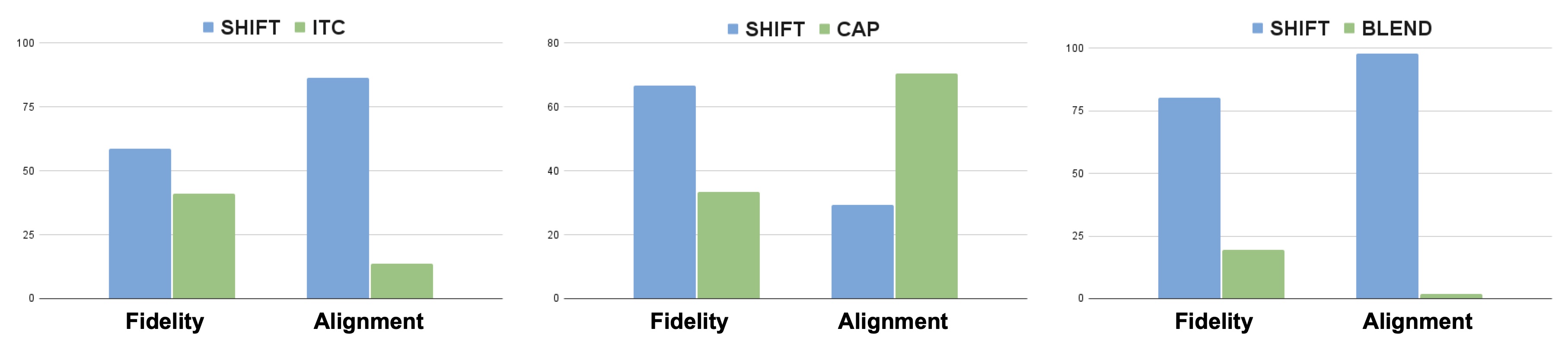}
  \vspace{-3mm}
  \caption{Human evaluation for photo-realism and condition-alignment.}
  \label{fig:fig3}
\end{figure}

Fig.\ref{fig:fig3} delivers the result. Compared to simple \textbf{ITC} baseline, \textbf{SHIFT} outperforms in both fidelity and alignment. Although \textbf{CAP}-only shows better condition alignment, \textbf{SHIFT} clearly outputs better quality samples, which is apparent from qualitative results as well. \textbf{BLEND}, a naive baseline that simply mixes \textbf{CAP} and \textbf{ITC} without gradual transition, performs significantly worse as these two signals can often be conflicting and difficult to optimize simultaneously.


\section{Conclusion}

In this paper, we have studied the semantic information encoded in different multimodal objectives by visually analyzing their properties as generative diffusion guidance. We hope it provides useful insights for ensuing works and spark\nj{s} further advances in the field.

\bibliography{iclr2023_conference}
\bibliographystyle{iclr2023_conference}

\newpage

\appendix
\section{Appendix}

\begin{figure}[h!]
\centering
  \includegraphics[width=0.94\textwidth]{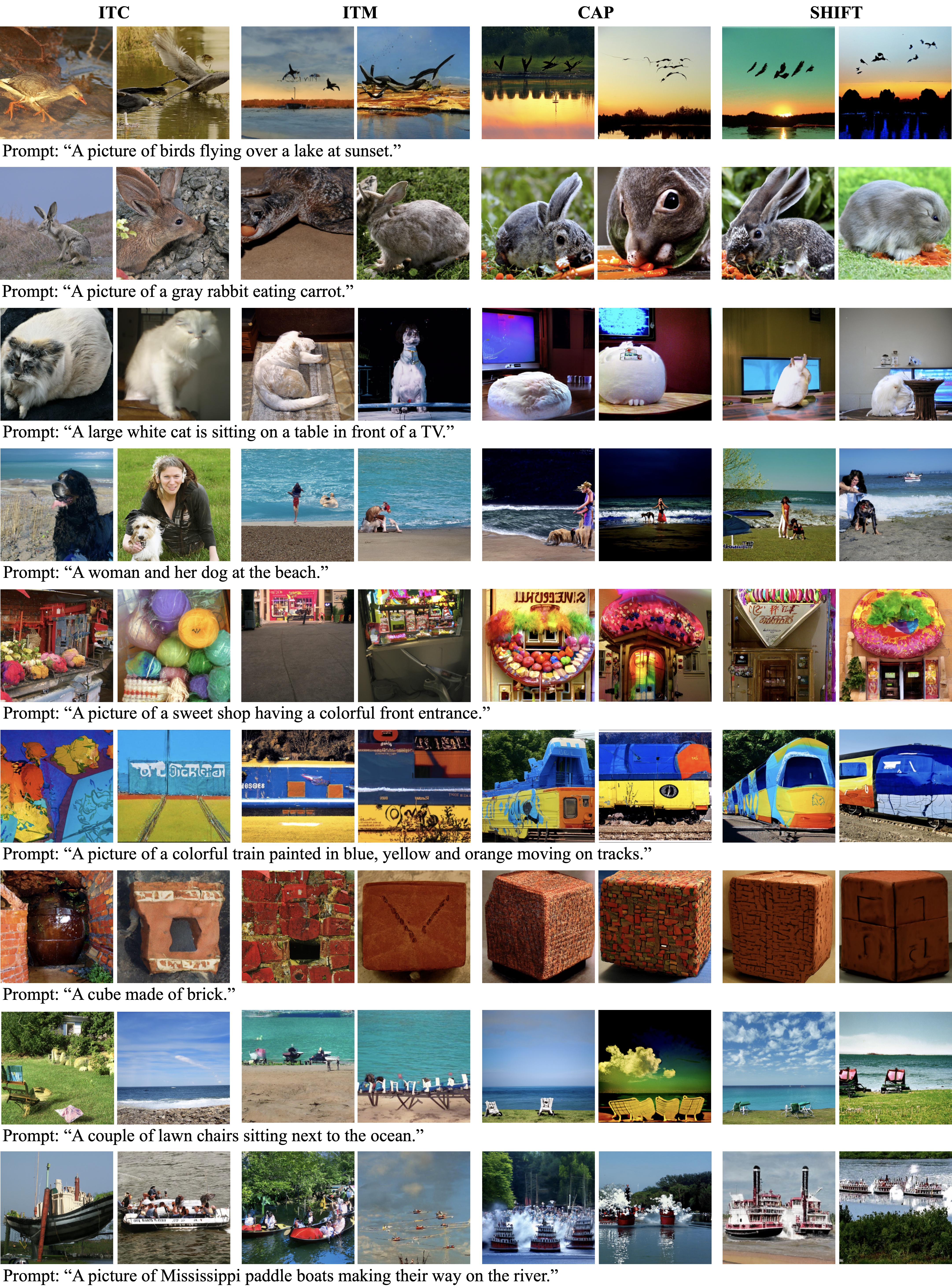}
  \vspace{-3mm}
  \caption{Additional text-to-image generation examples. We can consistently observe that while \textbf{ITC} focuses on detailed formulation of the salient object, \textbf{CAP} and its variant \textbf{SHIFT} understand the prompt in a finer level and output more faithful visualizations. It is also apparent that \textbf{ITC} alone often leaves out certain objects or mixes different visual semantics (\textit{e.g., colorful entrance}).}
  \label{fig:fig1}
\end{figure}

\begin{figure}[h!]
\centering
  \includegraphics[width=0.85\textwidth]{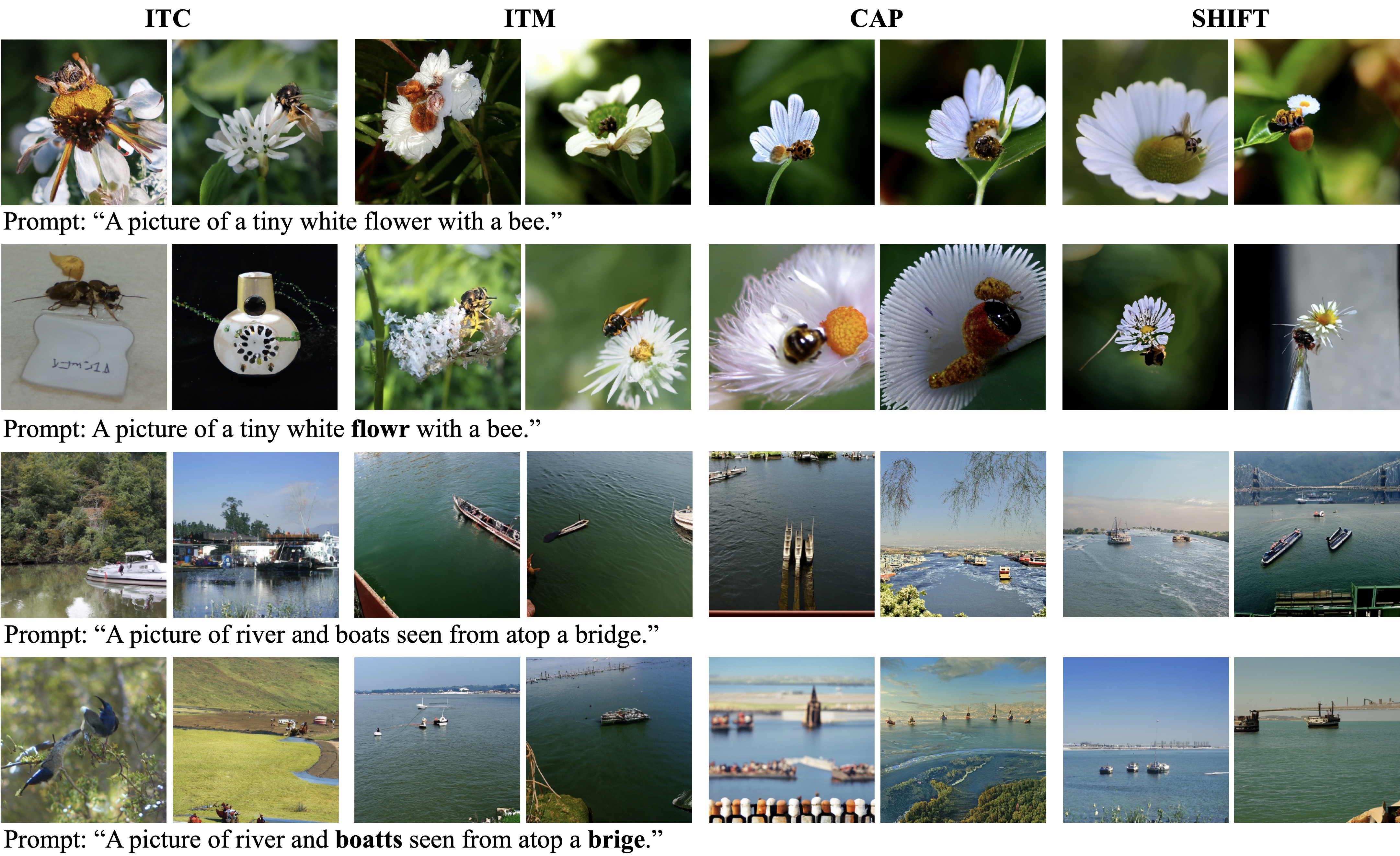}
  \vspace{-3mm}
  \caption{Additional experiments for noise robustness. Although \textbf{ITC} produces realistic images with clean prompts, minor typos can completely ruin their semantic signals. In contrast, losses that provide denser supervisions generally output consistent results despite textual noise, showing better robustness.}
  \label{fig:fig1}
\end{figure}

\begin{figure}[h!]
\centering
  \includegraphics[width=0.85\textwidth]{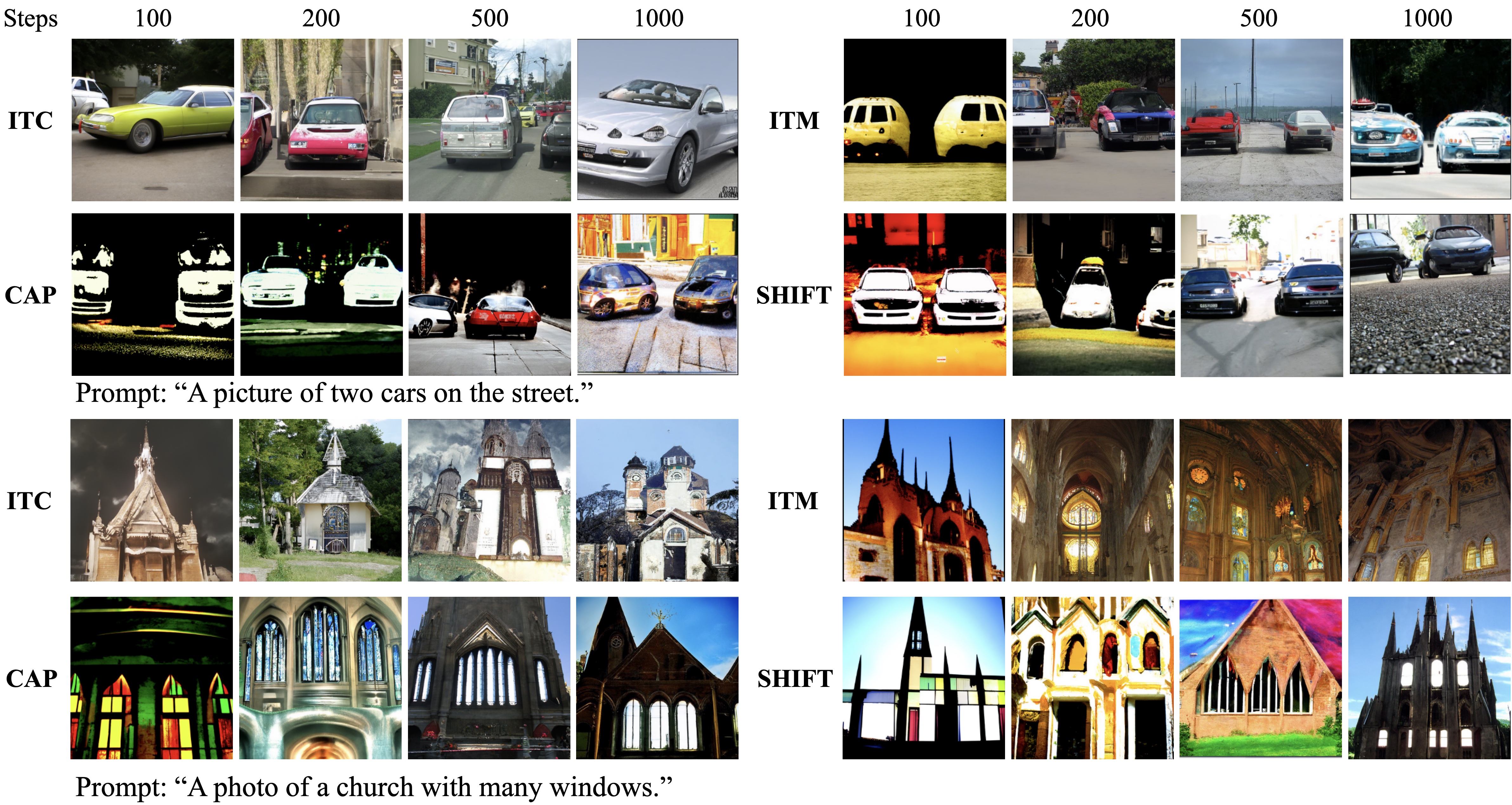}
  \vspace{-3mm}
  \caption{Additional results for optimization complexity. Captioning-based losses require more diffusion steps to generate realistic images, while \textbf{ITC} and \textbf{ITM} quickly forms reasonable shapes and appearances.}
  \label{fig:fig1}
\end{figure}

\begin{figure}[h]
\centering
  \includegraphics[width=0.85\textwidth]{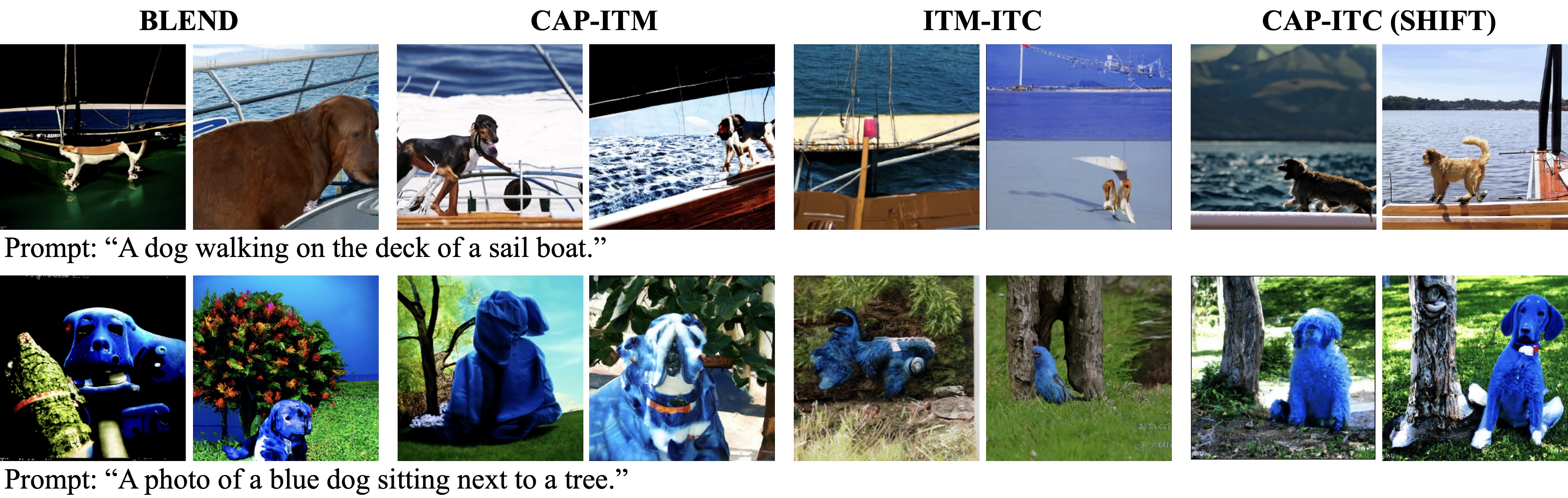}
  \vspace{-3mm}
  \caption{Qualitative comparison between baselines that combine multiple objectives. \textbf{BLEND} mixes \textbf{CAP} and \textbf{ITC} with no transition. We observe that gradually shifting from \textbf{CAP} to \textbf{ITC} enjoys advantages from both sides, \textit{i.e.,} faithful scene composition and realistic details.}
  \label{fig:fig1}
\end{figure}

\end{document}